\documentclass[lettersize,journal]{IEEEtran}
\usepackage{amsmath,amsfonts}

\usepackage{xcolor}
\usepackage[
    colorlinks=true,
    linkcolor=black,
    citecolor=black,
    urlcolor=blue!75!black
]{hyperref}

\usepackage{algorithmic}
\usepackage{algorithm}
\usepackage{array}
\usepackage[caption=false,font=normalsize,labelfont=sf,textfont=sf]{subfig}
\usepackage{textcomp}
\usepackage{stfloats}
\usepackage{url}

\usepackage{verbatim}
\usepackage{graphicx}
\usepackage{cite}
\usepackage{bm}
\usepackage{amsthm}
\usepackage{booktabs} 
\usepackage{makecell}
\usepackage{amssymb}
\usepackage{multirow} 

\begin{document}

\title{Multi-Turn Reasoning LLMs for Task Offloading in Mobile Edge Computing}

\author{Ning Yang,~\IEEEmembership{Senior Member,~IEEE}, {Chuangxin Cheng}, and {Haijun Zhang*,~\IEEEmembership{Fellow,~IEEE}}
\thanks{Ning Yang is with the Institute of Automation, Chinese Academy of Sciences, Beijing, 100190, China (e-mail: ning.yang@ia.ac.cn).}
\thanks{Chuangxin Cheng is with the School of Mechano-Electronic Engineering, Xidian University, Xi’an, 710071, China (e-mail: 25041212577@stu.xidian.edu.cn).}
\thanks{Haijun Zhang is with the School of Computer and Communication Engineering, University of Science and Technology Beijing, Beijing, 100083,
China (e-mail: zhanghaijun@ustb.edu.cn).}
\thanks{This work was supported by the National Natural Science Foundation of China under Grants 62301559.}
\thanks{(*Corresponding author: Haijun Zhang)}
}




\maketitle

\begin{abstract}
Emerging computation-intensive applications impose stringent latency requirements on resource-constrained mobile devices. Mobile Edge Computing (MEC) addresses this challenge through task offloading. However, designing effective policies remains difficult due to dynamic task arrivals, time-varying channels, and the spatio-temporal coupling of server queues. Conventional heuristics lack adaptability, while Deep Reinforcement Learning (DRL) suffers from limited generalization and architectural rigidity, requiring retraining when network topology changes. Although Large Language Models (LLMs) offer semantic reasoning capabilities, standard Supervised Fine-Tuning (SFT) yields myopic policies that greedily minimize immediate latency without accounting for long-term system evolution.
To address these limitations, we propose Collaborative Optimization via Multi-turn Large Language Models (COMLLM), a generative framework that enables foresighted decision-making in MEC systems. COMLLM integrates Group Relative Policy Optimization (GRPO) with a Look-Ahead Collaborative Simulation (LACS) mechanism, which performs multi-step Monte Carlo rollouts while jointly modeling server queue dynamics. By incorporating these rollouts into the reward design, the framework captures the long-term impact of current decisions on future system states.
Experimental results demonstrate that COMLLM achieves near-optimal latency and improved load-balancing fairness. Notably, it exhibits zero-shot topological scalability, allowing a model trained on small-scale networks to generalize to larger, unseen topologies without retraining, outperforming SFT, DRL, and heuristic baselines. The code is available at: \url{https://github.com/Cassandra-Cheng/COMLLM-MEC-Task-Offloading}.
\end{abstract}


\section{Introduction}
\subsection{Background and Motivation}
\IEEEPARstart{I}{n} recent years, the proliferation of smart mobile devices, such as drones, autonomous vehicles, and Internet of Things (IoT) sensors, has led to a surge in computation-intensive and latency-sensitive applications. Representative examples include augmented reality, real-time video analysis, and digital twin rendering~\cite{7879258, 8030322, 8016573}. These applications involve large-scale data processing and strict real-time constraints, thereby imposing stringent requirements on service quality, particularly in terms of ultra-low latency and high computational capability~\cite{10932256, 11417298}. However, mobile devices are inherently constrained by limited battery capacity, inadequate thermal management, and restricted local computing resources, making them incapable of supporting such demanding workloads~\cite{9712216, 10713973, 10163866}. Although offloading tasks to centralized cloud servers provides virtually unlimited computational resources, cloud computing architectures often fail to meet strict real-time requirements due to the propagation delay caused by long-distance data transmission, as well as potential congestion in the core network~\cite{7474412}.

Mobile Edge Computing (MEC) has emerged as an effective approach to overcome these limitations. By deploying computing, storage, and caching resources at the network edge in proximity to end users, MEC can effectively reduce service latency and alleviate the burden on the core network~\cite{7879258,yang2025minimizing}. However, achieving optimal task offloading decisions in dynamic MEC environments remains a fundamental challenge. This difficulty arises from three key characteristics of the system: random task arrivals, time-varying wireless channel conditions, and \textit{spatio-temporal coupling}~\cite{9733202}. In particular, spatio-temporal coupling implies that decisions made in the current time slot directly influence future queue states and resource availability, resulting in strong long-term dependencies in the decision-making process. These intertwined dynamics render traditional mathematical optimization and heuristic algorithms inadequate for real-time deployment, either due to prohibitive computational complexity or limited adaptability~\cite{10891825}.

To address such non-stationary and sequential decision-making problems, Deep Reinforcement Learning (DRL) has been adopted in MEC~\cite{9712216, 11144916}. By modeling the task offloading process as a Markov Decision Process (MDP), DRL is capable of optimizing long-term performance under uncertainty. However, its practical deployment in MEC systems is hindered by several intrinsic limitations, including the curse of dimensionality and limited generalization capability~\cite{9712216, 10713973}. Specifically, DRL policies operate on fixed-dimensional numerical state representations, making them sensitive to changes in network topology. In dynamic environments, where edge servers may be added or removed, such dimensional mismatch renders pre-trained policies incompatible. Consequently, adapting DRL to new network configurations often requires redesigning the model architecture and performing costly retraining, limiting its scalability across different network sizes.

Large language models (LLMs) offer a promising alternative to overcome these structural limitations~\cite{11417298, 11416846, khanna2025grl}. Unlike DRL, LLMs process variable-length, unstructured textual inputs, enabling a flexible representation of dynamic system states. By encoding numerical network information into natural language prompts, task offloading can be reformulated as a sequential reasoning problem. This representation supports varying numbers of servers, offering the potential for topology-agnostic generalization without architectural modifications. Furthermore, the contextual understanding capability of LLMs allows them to interpret complex system states as coherent sequences.

Nevertheless, existing LLM-based approaches, such as supervised fine-tuning (SFT) and in-context learning (ICL), remain limited in handling the long-term dependencies induced by spatio-temporal coupling~\cite{10400970, 11044664, 11039039, 10870187}. Since these methods primarily rely on imitation from historical data, they tend to exhibit myopic decision-making behaviors. In practice, such models often favor servers with higher computational capacity to minimize immediate latency, while neglecting the long-term impact on queue congestion. The resulting queue backlogs and increased task drop rates under bursty traffic conditions highlight the inability of imitation-based approaches to explicitly account for future system dynamics~\cite{11039039, 10870187, ZHU202511}.

In summary, an effective MEC task offloading framework should satisfy two essential requirements. First, it should achieve topology-agnostic generalization, so that it can adapt to varying numbers of edge servers without redesigning the policy architecture or retraining from scratch. Second, it should support foresighted decision-making by explicitly accounting for the long-term impact of current actions on future queue evolution and system congestion. Satisfying these two requirements simultaneously is nontrivial: the former demands structural flexibility in state representation, while the latter requires the policy to optimize delayed and implicitly manifested future effects. This motivates the development of a learning framework that not only leverages the representational flexibility of LLMs, but also explicitly incorporates future system dynamics into policy optimization during training.

\subsection{Solution Approach and Contributions}
To address the two core challenges discussed above, namely topology-agnostic generalization and foresighted decision-making in dynamic MEC environments, we propose \textit{Collaborative Optimization via Multi-turn Large Language Models} (COMLLM). COMLLM is an LLM-based task offloading framework that integrates semantic state representation, reinforcement fine-tuning, and future-aware reward shaping. Instead of treating the LLM as a static imitator of historical decisions, the proposed framework optimizes the policy directly in the MEC environment, enabling it to adapt to varying network topologies while accounting for the long-term impact of current offloading actions on future congestion and latency. In this framework, semantic state serialization supports variable-topology representation, GRPO serves as the policy optimization backbone, and the main technical novelty lies in the proposed LACS mechanism, which introduces physically grounded look-ahead simulation into the reward design to capture the impact of current actions on future queue congestion. The main contributions of this paper are summarized as follows:
\begin{itemize}
    \item \textbf{Future-Aware LLM Framework for MEC Task Offloading:} COMLLM is introduced as a task offloading framework that reformulates dynamic MEC decision-making as a language-conditioned sequential decision-making problem. Through semantic state serialization, it overcomes the fixed-dimensional input limitation of conventional DRL and provides a flexible solution for topology-varying MEC environments.

    \item \textbf{GRPO Training with Look-Ahead Collaborative Simulation:} Group Relative Policy Optimization (GRPO)~\cite{shao2024deepseekmath} is integrated with a novel Look-Ahead Collaborative Simulation (LACS) mechanism to mitigate the myopic behavior of existing SFT- or ICL-based LLM offloading methods. LACS performs look-ahead simulation within the reward loop to estimate the downstream congestion effect of candidate actions, enabling the policy to optimize not only immediate latency but also the long-term impact of current offloading decisions on queue evolution and server contention.

    \item \textbf{Zero-Shot Topology Transfer and Robust Decision-Making:} Through extensive experiments, we show that COMLLM generalizes effectively across unseen MEC topologies without architectural redesign or retraining, while also maintaining strong robustness under high-load and prompt-perturbed settings. The results demonstrate that COMLLM achieves superior latency, load balancing, and task completion performance compared with heuristic, DRL, and imitation-based LLM baselines.
\end{itemize}

\section{Related Work}
Existing MEC task offloading studies can be broadly divided into traditional optimization and heuristic methods, DRL-based approaches, and LLM-driven methods. We review these categories with emphasis on topology generalization and foresighted decision-making.

\textbf{Traditional Optimization and Heuristic Methods:}
Traditional studies formulate MEC task offloading as mathematical optimization, scheduling, or game-theoretic problems involving latency, energy, bandwidth, and computation constraints.

For joint offloading and resource allocation, Jiang \emph{et al.}~\cite{9712216} studied energy-constrained MEC and optimized offloading decisions together with resource allocation under coupled system constraints. Zhang \emph{et al.}~\cite{10737035} further investigated edge node allocation with user delay tolerance, aiming to reduce deployment cost while satisfying service requirements. For task dependency and task partitioning, Asheralieva \emph{et al.}~\cite{10696908} considered dependent delay-sensitive tasks in multi-operator multi-access networks, where the offloading decision must explicitly account for execution ordering and buffering effects. Gao \emph{et al.}~\cite{9542866} studied task partitioning and offloading in deep neural network (DNN)-enabled MEC networks, where the challenge lies in jointly deciding how tasks should be split and where each part should be processed. Beyond these general formulations, some works extended optimization and heuristic methods to specific MEC scenarios. Xu \emph{et al.}~\cite{10654497} developed a hybrid service selection strategy for unmanned aerial vehicle (UAV)-assisted MEC delivery systems, while Wu \emph{et al.}~\cite{10684062} and Park and Chung~\cite{9682175} explored architectural and collaborative offloading designs for IoT and edge collaboration settings.

Although analytically interpretable, these methods are generally tied to predefined variables and scenario-specific models, requiring re-formulation or re-optimization when network configurations change.

\textbf{DRL-based Approaches:}
To improve adaptability in dynamic MEC environments, DRL has been introduced into task offloading. By formulating offloading as a sequential decision-making problem, DRL enables policies to be learned directly through interaction with stochastic environments and is therefore more suitable for optimizing long-term objectives.

For long-term optimization under uncertainty, Tang and Wong~\cite{9253665} demonstrated the effectiveness of DRL for long-term task offloading optimization in MEC systems, showing that reinforcement learning can outperform myopic decision rules when future system dynamics matter. Huang \emph{et al.}~\cite{8771176} further extended this idea to wireless-powered MEC networks, where online DRL is used to handle dynamic energy harvesting and time-varying channels. For coordinated decision-making in complex scenarios, Ling \emph{et al.}~\cite{10696946} proposed a multi-agent DRL framework with an attention mechanism for MEC-enabled IoT, enabling coordinated offloading and resource allocation among multiple agents. Peng \emph{et al.}~\cite{10713973} studied sequential offloading in edge computing from a reinforcement perspective, while Xiao \emph{et al.}~\cite{9733202} considered dependent IoT applications in edge-intelligence systems, where task dependency further increases the difficulty of policy design. For enhanced state representation or optimization objectives, Chen \emph{et al.}~\cite{10568381} incorporated graph neural networks into DRL to better model topological dependencies in cooperative edge computing, while Yang \emph{et al.}~\cite{10349870,11144916} explored multi-objective and generalizable Pareto-optimal reinforcement learning to balance latency, energy consumption, fairness, and policy transferability.

However, most DRL policies rely on fixed-dimensional numerical states and remain coupled to the training topology. Changes in the number of servers therefore often require state-space redesign and model retraining.

\textbf{LLM-driven Task Offloading:}
Recently, LLMs have been explored as a new approach for MEC and cloud-edge offloading, mainly because they can process variable-length semantic inputs rather than fixed-dimensional numerical tensors. This makes them attractive for topology-varying MEC environments.

For semantic state serialization for topology flexibility, Song \emph{et al.}~\cite{10827049} directly explored task offloading with LLMs in MEC by converting system states into textual descriptions, showing the feasibility of using language models for offloading decisions. For SFT or ICL for decision generation, Zhou \emph{et al.}~\cite{10811953} studied generation task offloading in 6G edge-cloud systems through in-context learning, where the model leverages prompt examples rather than topology-specific network redesign. For LLM-assisted cloud-edge orchestration, He \emph{et al.}~\cite{10591707} considered LLM inference offloading and resource allocation in cloud-edge computing from an active inference perspective. Hu \emph{et al.}~\cite{10771985} proposed a cloud-edge collaborative architecture for multimodal LLM-based systems in IoT networks, while Jahan \emph{et al.}~\cite{11039039} further explored a generative-AI-based approach for computation offloading and resource management in collaborative vehicular MEC networks. In a broader sense, the recent survey on decision-making LLMs for wireless communication~\cite{11180008} also indicates that language-based decision frameworks support prompt engineering, semantic reasoning, and reward modeling for communication control tasks.

However, existing methods mainly rely on SFT, ICL, or high-level generative guidance and do not explicitly optimize the effect of current actions on future queue evolution. This limitation motivates the framework proposed in this paper, which aims to combine the structural flexibility of LLMs with explicit future-aware policy optimization for dynamic MEC task offloading.

\section{System Model and Problem Formulation}
In this section, we establish the physical model of the dynamic MEC environment and formulate the corresponding sequential offloading problem. The system architecture, user/task model, offloading decisions, and the communication and computation delay models are first described. Based on these physical constraints and queue evolution dynamics, the long-term optimization objective is then defined. For ease of reference, the core symbols introduced in this section are
summarized in Appendix~A.

\subsection{System Model}
We consider a representative MEC architecture consisting of a set of users and a set of heterogeneous edge servers deployed at the network edge. Let $\mathcal{U}=\{1,2,\dots,U\}$ denote the user set, where each user stochastically generates computation tasks, and let
$\mathcal{E}=\{1,2,\dots,E\}$ denote the set of available edge servers. In addition, define the set of candidate execution locations as $\tilde{\mathcal{E}}=\{0\}\cup\mathcal{E}$, where $e=0$ denotes local execution and $e\in\mathcal{E}$ denotes offloading to edge server $e$. To reflect realistic deployments, the edge servers may differ in computational capacities and real-time workload conditions. For example, a high-capacity server may still incur a longer delay when it has many active tasks and a large queue backlog, whereas a lower-capacity but lightly loaded server may provide faster service. The system evolves in a time-slotted manner. Let $\mathcal{T}=\{0,1,2,\dots\}$ denote the set of time slots, each with duration $\Delta t$. At the beginning of each step, the arrival time of a series of tasks follows a Poisson distribution for each user, and the Poisson arrival rate for each user is $\lambda_p$. Assume that the system state, including channel conditions and server workloads, remains quasi-static within each slot but may vary across slots. Let $\mathcal{M}=\{1,2,\dots,M\}$ denote the set of tasks in an episode.  Let task $m\in\mathcal{M}$ denote the current task under consideration. We define $\Phi_m =\{u_m,D_m,\gamma,\tau_m^{\max}\}$, where $u_m\in\mathcal{U}$ is the source user of task $m$, $D_m$ is the input data size, $\gamma$ is the computational density in CPU cycles per bit, and $\tau_m^{\max}$ is the maximum tolerable latency (deadline). Therefore, the total computational workload of task $\Phi_m$ is $D_m\gamma$ CPU cycles.


\textbf{Offloading Decision:} For each task $\Phi_m$, define a binary offloading vector $x_{m,e}\in\{0,1\},\forall e\in\tilde{\mathcal{E}}$, where $x_{m,e}=1$ indicates that task $m$ is assigned to edge server $e$. Since each task is assigned to exactly one execution location, the following constraint must hold: $\sum_{e\in\tilde{\mathcal{E}}} x_{m,e}=1.$ Consequently, the task latency is determined by the selected execution location: local execution follows the local computation model, while offloading to edge server $e\in\mathcal{E}$ follows the edge computation model associated with server $e$.

\textbf{Communication Delay Model:} When task $m$ is offloaded to edge server $e\in\mathcal{E}$, user $u_m$ needs to upload the input data $D_m$ through a dynamic wireless channel. We consider a block-fading channel model, under which the uplink transmission rate from user $u$ to edge server $e$ is
\begin{equation}
R_{u,e}^{\mathrm{up}} = B \log_2 \left(1+\frac{P|h_{u,e}|^2}{\sigma^2}\right),
\qquad \forall u\in\mathcal{U},\ \forall e\in\mathcal{E},
\end{equation}
where $B$ is the channel bandwidth, $P$ is the transmit power, $|h_{u,e}|^2$ is the channel power gain, and $\sigma^2$ is the noise power. The uplink transmission delay of task $m$ is
\begin{equation}
T_m^{\mathrm{up}}=\sum_{e\in\mathcal{E}} x_{m,e}\frac{D_m}{R_{u_{m,e}}^{\mathrm{up}}},\qquad \forall m\in\mathcal{M}.
\end{equation}
Here, the summation is not used to accumulate transmission delays across multiple edge servers. Instead, it serves as a selection operator under the binary offloading decision variables. Since each task is assigned to only one execution location, if task $m$ is offloaded to edge server $e^\star$, then $x_{m,e^\star}=1$ and $x_{m,e}=0$ for all $e\neq e^\star$, yielding $T_m^{\mathrm{up}}=D_m/R_{u_m,e^\star}^{\mathrm{up}}$. If task $m$ is executed locally, then $x_{m,e}=0$ for all $e\in\mathcal{E}$ and no uplink transmission delay is incurred.

\textbf{Local Computation Delay Model:} If task $m$ is executed locally, the corresponding local execution latency is
\begin{equation}
T_{m,0}^{\mathrm{loc}} = T_{m,0}^{\mathrm{wait,loc}} + \frac{D_m\gamma}{f^{\mathrm{loc}}},\qquad e=0,\forall m\in\mathcal{M},
\end{equation}
where $f^{\mathrm{loc}}$ is the local CPU frequency and $T_{m,0}^{\mathrm{wait,loc}}$ represents the local waiting delay induced by the residual workload already buffered at the mobile device, and is treated as part of the observable system state. This model captures the fact that, although local execution avoids communication overhead, it is limited by the relatively low computation capability of the mobile device.

\textbf{Edge Computation Delay Model:} If the task is offloaded to edge server $e$, it experiences uplink transmission, queueing delay, and execution delay. Let $F_e$ denote the maximum computation capacity of edge server $e$, and let $N_e^m$ denote the number of active tasks already sharing edge server $e$ when task $m$ arrives. Under a processor-sharing (PS) model, the effective computation rate allocated to the arrived task is
\begin{equation}
f_{m,e}^{\mathrm{eff}} = \frac{F_e}{N_e^m+1},
\qquad \forall e\in\mathcal{E},\forall m\in\mathcal{M},
\end{equation}
where the term $+1$ accounts for the offloaded task itself. The corresponding execution time on edge server $e$ is
\begin{equation}
T_{m,e}^{\mathrm{exec}} = \frac{D_m\gamma}{f_{m,e}^{\mathrm{eff}}}
= \frac{D_m\gamma(N_e^m+1)}{F_e},\forall e\in\mathcal{E},\forall m\in\mathcal{M}.
\end{equation}
In addition, let $L_e^m$ denote the workload backlog of edge server $e$ when task $m$ arrives, measured in bits. Here, $N_e^m$ characterizes the instantaneous degree of processor sharing at edge server $e$, while $L_e^m$ captures the remaining queued workload awaiting service. Then the queueing delay experienced by the new task is approximated by
\begin{equation}
T_{m,e}^{\mathrm{wait}} = \frac{L_e^m\gamma}{f_{m,e}^{\mathrm{eff}}},
\qquad \forall e\in\mathcal{E},\forall m\in\mathcal{M}.
\end{equation}
Therefore, when task $m$ is offloaded to edge server $e$, the total edge latency is
\begin{equation}
T_{m,e}^{\mathrm{edge}} = \frac{D_m}{R_{u_{m,e}}^{\mathrm{up}}} + T_{m,e}^{\mathrm{wait}} + T_{m,e}^{\mathrm{exec}},
\qquad \forall e\in\mathcal{E},\forall m\in\mathcal{M}.
\end{equation}
For notational convenience, define the candidate execution latency as
\begin{equation}
T_{m,e}=
\begin{cases}
T_{m,0}^{\mathrm{loc}}, & e=0,\forall m\in\mathcal{M},\\
T_{m,e}^{\mathrm{edge}}, & e\in\mathcal{E},\forall m\in\mathcal{M}.
\end{cases}
\end{equation}

\textbf{Queue Evolution Dynamics:} To capture temporal coupling across tasks, we model the workload evolution of each edge server queue. The equivalent amount of input data that can be processed under the effective service rate of edge server $e$ during one slot is
\begin{equation}
R_{m,e}^{\mathrm{process}} = \frac{f_{m,e}^{\mathrm{eff}}\Delta t}{\gamma},
\qquad \forall e\in\mathcal{E},\forall m\in\mathcal{M}.
\end{equation}
Consequently, the backlog at edge server $e$ evolves as
\begin{equation}
L_e^{m+1} = \max\left(0, L_e^m - R_{m,e}^{\mathrm{process}}\right) + D_m x_{m,e},
\end{equation}
which shows that the current offloading decision directly affects future queue states, thereby inducing strong temporal coupling in the decision process.

\subsection{Problem Formulation}
Based on the above system model, the objective is to design an offloading policy that minimizes the long-term system cost while satisfying task deadlines as much as possible. A stochastic offloading policy is defined as $\pi:\mathcal{S}\times\mathcal{A}\rightarrow [0,1]$, where $\mathcal{S}$ is the system state space and $\mathcal{A}$ is the offloading action space, both of which will be formally defined in Section~IV. For a given task $m$ and observable system state, the policy $\pi$ selects an offloading decision vector $\mathbf{x}_m=\{x_{m,e}\}_{e\in\tilde{\mathcal{E}}}$ according to a probability distribution. For task $m$, the latency induced by the offloading decision is defined as $C_m=\sum_{e\in\tilde{\mathcal{E}}}x_{m,e}T_{m,e}$. To account for deadline violations, we introduce a penalty term $\rho$ and define the generalized task cost as
\begin{equation}
J_m=C_m+\mathbb{I}\!\left(C_m>\tau_m^{\max}\right)\rho,
\end{equation}
where the indicator term penalizes any decision whose resulting latency exceeds the deadline $\tau_m^{\max}$. Then the sequential offloading problem can be formulated as
\begin{subequations}
\begin{align}
\min_{\pi} \quad & \mathbb{E}_{\mathbf{x}\sim \pi}\!\left[\sum_{m\in\mathcal{M}} J_m(\mathbf{x}_m)\right] \label{eq:p1_obj}\\
\text{s.t.} \quad & x_{m,e} \in \{0,1\}, \quad \forall m\in\mathcal{M},\ \forall e\in\tilde{\mathcal{E}}, \label{eq:p1_c1}\\
& \sum_{e\in\tilde{\mathcal{E}}} x_{m,e}=1, \quad \forall m\in\mathcal{M}, \label{eq:p1_c2}\\
& N_e^m \ge 0,\ \ L_e^m \ge 0, \quad \forall e\in\mathcal{E},\ \forall m\in\mathcal{M}, \label{eq:p1_c3}\\
& L_e^{m+1} = \max\left(0, L_e^m-\frac{f_{m,e}^{\mathrm{eff}}\Delta t}{\gamma}\right)+D_mx_{m,e}. \label{eq:p1_c4}
\end{align}
\end{subequations}

In this problem, \eqref{eq:p1_obj} minimizes the expected cumulative generalized cost induced by the stochastic offloading policy. Constraint \eqref{eq:p1_c1} specifies the binary decision structure. Constraint \eqref{eq:p1_c2} ensures that each task is assigned to exactly one execution location. Constraint \eqref{eq:p1_c3} guarantees the non-negativity of the queue-related variables. Constraint \eqref{eq:p1_c4} describes the queue evolution of each edge server, indicating that the current offloading decision affects not only the immediate latency but also future server congestion. The problem is difficult to solve for three reasons. First, the decision variables are discrete, while the delay and queue dynamics are nonlinear, which makes the problem a sequential mixed discrete optimization problem. Second, the state space grows rapidly with the number of heterogeneous edge servers, rendering dynamic programming intractable in real time. Third, the queue evolution constraint in \eqref{eq:p1_c4} couples current decisions with future system states, so a locally optimal action may lead to poor long-term performance.

\section{Methodology}
To solve the optimization problem in real time, we reformulate the physical MEC environment into a MDP and propose the \textbf{COMLLM} framework, which integrates LLMs with reinforcement learning to learn a high-quality offloading policy.

\subsection{MDP Formulation}
For the task under consideration at decision step \(t\), the binary offloading decision vector \(\{x_{m,e}\}\) defined in Section~III induces the discrete action $a_t=\sum_{e\in\tilde{\mathcal{E}}} e\,x_{m,e}$, where \(a_t=0\) denotes local execution and \(a_t=e\in\mathcal{E}\) denotes offloading to edge server \(e\). Based on this correspondence between the physical offloading decision and the MDP action, the subsequent methodology is presented in the standard MDP form using \(s_t\), \(a_t\), and \(r_t\). The physical MEC system defined in Section~III is then reformulated as an MDP tuple \(\langle \mathcal{S}, \mathcal{A}, \mathcal{P}_{\mathrm{trans}}, \mathcal{R} \rangle\) to enable policy learning.

\textbf{State Space $\mathcal{S}$:}
At each decision step $t$, the state $s_t \in \mathcal{S}$ represents the observable MEC environment:
\begin{equation}
\begin{aligned}
s_t &= \Big\{ \Phi_m,\ \{R_{u_{m,e}}^{\mathrm{up}}\}_{e \in \mathcal{E}},\ f^{\mathrm{loc}},\ T_{m,0}^{\mathrm{wait,loc}}, \\
&\qquad \{F_e, N_e^m, L_e^m, \mathbf{H}_e^m\}_{e \in \mathcal{E}} \Big\},
\end{aligned}
\end{equation}
where $\mathbf{H}_e^m=[h_e^{m-H},\ldots,h_e^{m-1}]$ denotes the workload history of edge server $e$ over the previous $H$ task-arrival instants available when task $m$ is considered. Here, $h_e^j$ denotes the workload observation of server $e$ associated with task $j$. It is introduced as an auxiliary semantic feature for the LLM, rather than as a required variable for the physical Markov property.

\textbf{Action Space $\mathcal{A}$:}
The action space is defined as $\mathcal{A}=\{0,1,\dots,E\}$, where $a_t=0$ indicates local execution and $a_t=e \in \mathcal{E}$ indicates offloading to edge server $e$. The LLM policy $\pi_\theta(a_t \mid s_t)$ outputs the discrete action token conditioned on the semantic state prompt. The generated response is deterministically parsed into a discrete action before interacting with the MEC environment. A response that cannot be mapped to local execution or an available edge server in $\mathcal{A}$ is treated as invalid and assigned a negative reward during policy optimization; thus, free-form text is not directly used as an executable MEC decision.

\textbf{Transition Kernel $\mathcal{P}_{\mathrm{trans}}$:}
The state transition probability $\mathcal{P}_{\mathrm{trans}}(s_{t+1}\mid s_t,a_t)$ is induced by the stochastic wireless channel evolution, random task arrivals, local residual workload, and the edge-server queue dynamics defined in Section~III. The queue evolution constraint in \eqref{eq:p1_c4} makes the transition explicitly action-dependent, thereby creating temporal coupling between current decisions and future congestion states.

\textbf{Reward Function $\mathcal{R}$:}
The original objective in \eqref{eq:p1_obj} is to minimize the cumulative generalized cost. Accordingly, we define the base one-step reward as the negative of the generalized physical cost:
\begin{equation}
r_t^{\mathrm{base}} = -J_t(a_t),
\label{eq:base_reward}
\end{equation}
where $J_t(a_t)$ denotes the step-wise cost induced by the selected action $a_t$, corresponding to $J_m$ under the mapping $m=t$. However, optimizing only the immediate reward is insufficient for future-aware offloading, which motivates the look-ahead reward shaping mechanism introduced in Section~\ref{subsec:lacs}.

\subsection{The COMLLM Framework Overview}
Unlike existing LLM-based offloading methods that mainly rely on prompt-based decision generation or imitation-style fine-tuning, COMLLM explicitly combines semantic generalization with reinforcement learning and look-ahead reward shaping for long-term policy optimization. COMLLM first converts heterogeneous MEC states into semantically structured prompts, allowing one LLM policy to process variable-size server sets without architectural redesign. On top of this semantic state abstraction, COMLLM uses SFT to initialize the policy from oracle-labeled data, then applies GRPO to improve the policy directly through interaction with the MEC environment. Finally, to address the short-sightedness of one-step reward optimization, COMLLM introduces LACS, which estimates the downstream congestion impact of the current action and injects this information into the reward signal.

\begin{figure*}[!t]
    \centering
    \includegraphics[width=0.95\textwidth]{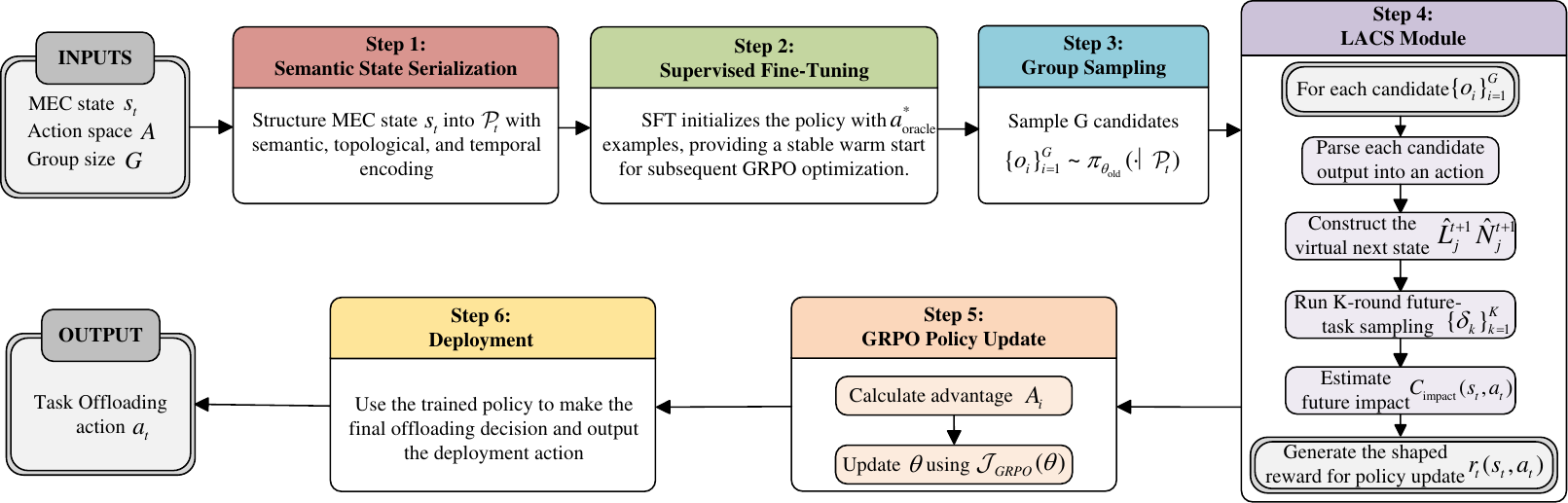}
    \caption{Overview of COMLLM. MEC states are serialized into textual prompts for SFT initialization. During GRPO training, candidate actions are evaluated using the physical MEC cost and LACS, whereas deployment requires only a single policy inference.}
    \label{fig_1}
\end{figure*}

\subsection{Supervised Fine-Tuning}
While pre-trained LLMs exhibit strong general-purpose reasoning, they lack the domain-specific expertise required for MEC, such as understanding network latency constraints and server queuing dynamics. We therefore design an SFT stage to bridge the gap between raw physical states and language-model-based decision generation.

The raw system state $s_t$ contains heterogeneous numerical variables with different scales and physical meanings. To make these variables more accessible to the LLM, we serialize them into structured text prompts through three principles:
\begin{itemize}
    \item \textbf{Semantic Unit Tagging:} Numerical values are explicitly annotated with physical units, such as GHz, Mbits, or Mbps, so that the LLM can better associate them with computation capacity, workload size, and channel quality.
    \item \textbf{Dynamic Topology Abstraction:} Edge servers are serialized as a variable-length list of key-value descriptions rather than a fixed-dimensional tensor, enabling the same model to handle unseen numbers of servers.
    \item \textbf{Temporal Load Encoding:} The recent workload history of each server is incorporated as a sequence field, allowing the LLM to capture short-term congestion trends.
\end{itemize}
This serialization is the key mechanism by which COMLLM acquires topology-agnostic generalization: the decision policy no longer depends on a fixed input dimension, but instead reasons over a semantically structured and variable-length representation. A complete prompt template, translated from the Chinese prompt used during training, together with its design rationale, is provided in Appendix~A.

To provide the SFT model with high-quality supervision, we construct an offline instruction-following dataset using an oracle. For each sampled system state $s_t$, the oracle evaluates all feasible actions in $\mathcal{A}$ according to the physical cost model in Section~III and selects the one-step optimal action:
\begin{equation}
a_{\mathrm{oracle}}^{*} = \arg\min_{a_t \in \mathcal{A}} J_t(a_t).
\end{equation}
The numerical optimal action $a_{\mathrm{oracle}}^{*}$ is then mapped to its corresponding natural language decision string, which serves as the ground truth response $y$. Through the aforementioned pipeline, an instruction-following dataset $\mathcal{D}_{SFT} = \{(x_i, y_i)\}_{i=1}^{N}$ is constructed. Here, $x_i$ denotes the tokenized input prompt that encapsulates the system state and task instruction, while $y_i=(y_{i,1},y_{i,2},\dots,y_{i,L_i})$ denotes the target decision sequence of length $L_i$. The parameters $\theta$ of the LLM are fine-tuned by minimizing the standard negative log-likelihood loss (i.e., cross-entropy) over the dataset $\mathcal{D}_{SFT}$:
\begin{equation}
\mathcal{L}_{SFT}(\theta) = - \mathbb{E}_{(x, y) \sim \mathcal{D}_{SFT}} \left[ \sum_{l=1}^{|y|} \log \pi_\theta(y_l | x, y_{<l}) \right],
\end{equation}
where $\pi_\theta(y_l | x, y_{<l})$ denotes the probability of generating the $l$-th token $y_l$, conditioned on the input prompt $x$ and the preceding generated tokens $y_{<l}$. This phase serves as a critical warm start, initializing the policy with prior domain knowledge and reducing random exploration at the beginning of reinforcement learning. It enables the model to learn the syntactic format of MEC decisions and the basic correlation between server specifications and latency, thereby providing an initialization for the subsequent GRPO phase.

\subsection{Group Relative Policy Optimization}
While SFT provides an initialization, it remains fundamentally imitation-based and cannot adapt to unseen states and long-horizon decision trade-offs. Therefore, the policy is further optimized using \textbf{GRPO}, a reinforcement learning method suited to LLM-based reasoning tasks because it avoids a separate critic network and instead estimates relative advantage from grouped samples~\cite{shao2024deepseekmath}.

\subsubsection{Group Sampling and Advantage Estimation}
Given a state prompt $\mathcal P_t$, the sampling policy $\pi_{\theta_{\mathrm{old}}}$ generates a group of $G$ candidate outputs:
\begin{equation}
\{o_i\}_{i=1}^{G} \sim \pi_{\theta_{\mathrm{old}}}(\cdot \mid \mathcal P_t).
\end{equation}
Each output is parsed into a discrete offloading
action. An output that cannot be mapped to a feasible action in
$\mathcal{A}$ is regarded as invalid and assigned a negative
reward, whereas a valid action is evaluated using the physical cost model
and the LACS-shaped reward defined in Section~\ref{subsec:lacs}. The
resulting scalar reward is denoted by $r_i$. Normalize the group rewards to obtain a relative advantage:
\begin{equation}
A_i = \frac{r_i-\mu_{\mathrm{rew}}}{\sigma_{\mathrm{rew}}+\epsilon_{\mathrm{adv}}},
\end{equation}
where $\mu_{\mathrm{rew}}, \sigma_{\mathrm{rew}}$ are the mean and standard deviation of rewards within the group. This group relative mechanism acts as a dynamic baseline, encouraging the model to increase the probability of actions that outperform their peers in the same batch.

\subsubsection{GRPO Objective}
To optimize the policy while controlling update instability and retain prior knowledge, let $\pi_{ref}$ represent the frozen reference policy strictly initialized from the model obtained in the SFT phase. The GRPO objective is defined as:
\begin{equation}\label{eq:grpo_objective}
\begin{aligned}
\mathcal{J}_{GRPO}(\theta) = \mathbb{E}_{T} \Bigg[ \frac{1}{G} \sum_{i=1}^{G} \bigg( \min \Big( \rho_i A_i, & \\
\text{clip}(\rho_i, 1-\epsilon, 1+\epsilon) A_i \Big) - \beta \mathbb{D}_{KL}&(\pi_\theta || \pi_{ref}) \bigg) \Bigg].
\end{aligned}
\end{equation}
Here, $\rho_i = \frac{\pi_\theta(o_i|\mathcal P_t)}{\pi_{\theta_{old}}(o_i|\mathcal P_t)}$ denotes the probability ratio between the current policy and the old sampling policy. Clipping limits excessively large updates relative to
$\pi_{\theta_{\mathrm{old}}}$, whereas the KL penalty
$\mathbb{D}_{\mathrm{KL}}(\pi_\theta\|\pi_{\mathrm{ref}})$ constrains deviation from the frozen SFT-initialized reference policy, helping preserve semantic knowledge and output-format correctness. A formal policy-improvement lower bound and its proof are provided in Appendix~A.

\subsection{Multi-turn Aware Reward Mechanism via Collaborative Simulation}
\label{subsec:lacs}
A fundamental challenge in MEC task offloading is the \emph{spatio-temporal coupling} of resource allocation: the action chosen at slot $t$ changes future queue states and thus affects the quality of subsequent decisions. Therefore, relying only on the one-step reward $r_t^{\mathrm{base}}(a_t)=-J_t(a_t)$ is insufficient for learning a truly future-aware policy. To address this issue, we propose a \textbf{LACS} mechanism, which estimates the downstream congestion effect of the current action by simulating a virtual transition and evaluating the residual system capability under sampled future tasks. For a current state $s_t$ and candidate action $a_t=e$, LACS performs the following three steps.

\textbf{Virtual State Transition:}
a virtual next state $\hat{s}_{t+1}$ is constructed by applying the physical queue transition defined in Section~III. For the selected server $e\in\mathcal{E}$, we update
\begin{equation}
\hat{L}_{e}^{t+1}
=
\max\left(0,L_e^t-\frac{f_{t,e}^{\mathrm{eff}}\Delta t}{\gamma}\right)+D_t,
\hat{N}_{e}^{t+1}=N_e^t+1.
\end{equation}
For all other servers $e'\neq e$, the virtual states remain unchanged except for their normal service evolution. This virtual transition captures how the current offloading action consumes future service capacity. When $a_t=0$, the task is executed locally and the edge-server queues evolve according to their normal service dynamics.

\textbf{Stochastic Future Task Sampling:}
To estimate the residual capability of the virtualized system, we sample $K$ future tasks $\{\delta_k\}_{k=1}^{K}$ by Monte Carlo simulation. For simplicity, the task size is sampled around the current workload scale:
\begin{equation}
\mathrm{Size}(\delta_k)\sim {D_{size}}(0.5D_t,\,1.5D_t), \qquad k=1,\dots,K.
\end{equation}
This stochastic sampling provides a tractable approximation of possible near-future traffic conditions.

\textbf{Oracle Evaluation of Residual Capability:}
For each sampled future task $\delta_k$, we evaluate its best cost under the virtual next state $\hat{s}_{t+1}$. Specifically, let $\tilde{C}_{t+1}^{(k)}(a \mid \hat{s}_{t+1})$ denote the cost of serving $\delta_k$ with action $a \in \mathcal{A}$ under the simulated state $\hat{s}_{t+1}$. Then the future cost for task $\delta_k$ is defined as
\begin{equation}
C_{\mathrm{future}}^{(k)}(\hat{s}_{t+1})
=
\min_{a \in \mathcal{A}} \tilde{C}_{t+1}^{(k)}(a \mid \hat{s}_{t+1}).
\end{equation}
Averaging over the $K$ sampled future tasks yields the future-impact term
\begin{equation}
C_{\mathrm{impact}}(s_t,a_t)
=
\frac{1}{K}\sum_{k=1}^{K}
C_{\mathrm{future}}^{(k)}(\hat{s}_{t+1}),
\end{equation}
where $K$ is the number of sampled future tasks. A larger $C_{\mathrm{impact}}(s_t,a_t)$ indicates that the current action leaves the system in a less favorable state for serving subsequent tasks. Based on this simulation, we define the LACS-shaped reward as
\begin{equation}
r_t(s_t,a_t)
=
-\Big(
J_t(a_t)
+\lambda\, C_{\mathrm{impact}}(s_t,a_t)
\Big),
\label{eq:lacs_reward}
\end{equation}
where $J_t(a_t)$ is the cost of the current task defined in Section~III, and $\lambda \ge 0$ controls the trade-off between immediate cost minimization and future congestion awareness. Using the reward in \eqref{eq:lacs_reward}, GRPO optimizes the discounted return
\begin{equation}
G_t^{\mathrm{LACS}}
=
\sum_{\ell=0}^{\infty}
\eta^\ell r_{t+\ell}(s_{t+\ell},a_{t+\ell}),
\end{equation}
where $\eta \in (0,1)$ is the discount factor, and $\ell$ is the time-offset index that counts how many steps ahead the reward is measured relative to the current slot $t$. In this way, LACS does not change the discounted-return formulation itself; instead, it reshapes each one-step reward so that the cumulative return better reflects the long-term congestion effect of current offloading decisions.

Therefore, LACS serves as a tractable surrogate for multi-step future risk. In practice, it discourages the policy from repeatedly selecting the currently strongest server when doing so would degrade the service capability available to subsequent tasks. The theoretical analysis of the LACS-induced future-cost approximation is provided in Appendix~A.

\section{Performance Evaluation}
In this section, we conduct extensive simulations to evaluate the effectiveness of the proposed COMLLM framework. COMLLM is compared with representative reinforcement learning and LLM-based baselines under dynamic MEC environments.

\begin{table}[htbp]
\centering
\caption{Parameter settings in the experimental setup}
\label{tab:parameter_settings}
\renewcommand{\arraystretch}{1.15}
\setlength{\tabcolsep}{6pt}
\begin{tabular}{lll}
\hline
\textbf{Category} & \textbf{Parameter} & \textbf{Value} \\
\hline
\multirow{8}{*}{\makecell[l]{System \\ Environment}}
& Time slot duration $\Delta t$ & 0.1 s \\
& Local CPU frequency $f^{\mathrm{loc}}$ & 2 GHz \\
& Edge server CPU frequency $F_e$ & $[20.0,48.0]$ GHz \\
& Average uplink rate $R_{u,e}^{\mathrm{up}}$ & 14 Mbps \\
& Task size $D_t$ & $[2.0, 5.0]$ Mbits \\
& Computational density $\gamma$ & 0.297 gigacycles/Mbit \\
& Task deadline $\tau_t^{\max}$ & 10 time slots (1 s) \\
& Task arrival probability $\lambda_p$ & 0.3 \\
\hline
\multirow{2}{*}{LACS}
& Look-ahead steps $K$ & 3 \\
& Future task size sampling & $D_{size}(0.5D_t,\,1.5D_t)$ \\
\hline
\multirow{2}{*}{\makecell[l]{Reward \\ Weights}}
& LACS reward weight $\lambda$ & 0.3 \\
& Deadline penalty $\rho$ & 10 \\
\hline
\end{tabular}
\end{table}

\subsection{Experimental Setup}
\textbf{Environment Configuration:}
We simulate a dynamic MEC environment following the system model in Section~III. The default environment contains 6 edge servers. The edge servers are heterogeneous, with their computation capacities sampled from the range shown in Table~\ref{tab:parameter_settings}. The uplink transmission condition is dynamically generated according to the wireless model in Section~III, with an average transmission rate of 14 Mbps. Task arrival probability is 0.3. Task size and future task size are randomly generated according to the parameter ranges in Table~\ref{tab:parameter_settings}.

\textbf{Dataset Construction:}
To comprehensively evaluate the models' performance and generalization capability, we generate three distinct datasets:

\begin{itemize}
    \item \textbf{SFT Dataset:} A training set of 1,000 samples is generated for SFT. Each sample consists of a randomly sampled MEC state and its corresponding oracle action label.
    \item \textbf{GRPO Dataset:} A larger interaction dataset of 2,000 samples is generated for reinforcement learning with GRPO.
    \item \textbf{Test Dataset:} A held-out set of 1,000 samples is independently generated for evaluation only. It follows the same server CPU range, task-size distribution, and task-arrival process as the training datasets, while the specific task instances and system states are unseen during training.
\end{itemize}
All compared methods are evaluated on the same test set for fair comparison. The key hardware configurations, training parameters, and computational costs of the SFT and GRPO stages are provided in Appendix~A.

\textbf{Evaluation Metrics:}
To comprehensively evaluate offloading quality, deadline satisfaction, performance efficiency, and workload distribution, we adopt four metrics: Average Latency, Task Drop Rate, Performance Ratio, and Load Balancing Index.

Average Latency measures the mean service cost achieved by a method over the test set. Since the generalized physical cost in Section~III is used as the per-task evaluation cost in our implementation, the average latency is computed as
\begin{equation}
\mathrm{AL}
=
\frac{1}{N}\sum_{t=1}^{N} J_t(a_t),
\label{eq:avg_latency_metric}
\end{equation}
where $N$ is the number of test samples and $J_t(a_t)$ denotes the generalized task cost incurred by action $a_t$. For tasks completed within their deadlines, $J_t(a_t)$ equals the actual service latency; for deadline-violating tasks, it additionally includes the penalty $\rho$.

Task Drop Rate quantifies the fraction of tasks whose realized latency exceeds the maximum tolerable delay. It is defined as
\begin{equation}
\mathrm{TDR}~(\%)
=
\frac{100}{N}\sum_{t=1}^{N}
\mathbb{I}\!\left(J_t(a_t)>\tau_t^{\max}\right),
\label{eq:tdr_metric}
\end{equation}
where $\tau_t^{\max}$ is the latency deadline.

To normalize model performance against a strong reference, define the Performance Ratio as
\begin{equation}
\mathrm{PR}~(\%)
=100\times
\frac{\bar{C}^{\mathrm{oracle}}}{\mathrm{AL}},
\label{eq:pr_metric}
\end{equation}
where $\mathrm{AL}$ is the average latency achieved and
\begin{equation}
C_t^{\mathrm{oracle}}
=
\min_{a_t\in\mathcal{A}} J_t(a_t),
\qquad
\bar{C}^{\mathrm{oracle}}
=
\frac{1}{N}\sum_{t=1}^{N} C_t^{\mathrm{oracle}}.
\end{equation}
That is, the oracle reference is constructed by exhaustive one-step action evaluation under the physical cost model. It is used only as a per-state reference for normalizing instantaneous decision quality and should not be interpreted as a globally optimal long-horizon policy.

To quantify how evenly the offloaded tasks are distributed across edge servers, we use Jain's Fairness Index as the Load Balancing Index. Let
\begin{equation}
n_e
=
\sum_{t=1}^{N}\mathbb{I}(a_t=e),
\qquad e=1,\ldots,E,
\end{equation}
where $n_e$ denotes the number of tasks assigned to edge server $e$ over the evaluation period. The Load Balancing Index is defined as
\begin{equation}
\mathrm{LBI}~(\%)
=
100\times
\frac{\left(\sum_{e=1}^{E}n_e\right)^2}
{E\sum_{e=1}^{E}n_e^2}.
\label{eq:lbi_metric}
\end{equation}
Local execution is excluded from the calculation because the metric evaluates workload distribution among edge servers. A higher LBI indicates a more balanced allocation, with $\mathrm{LBI}=100\%$ representing perfectly uniform task distribution across all edge servers.

\textbf{Comparison Methods:}
We compare COMLLM against the following representative baselines:
\begin{itemize}
    \item \textbf{Random:} A policy that uniformly samples an action from the feasible action space $\mathcal{A}=\{0,1,\dots,E\}$.
    \item \textbf{Deep Q-Network (DQN):} A representative value-based DRL baseline.
    \item \textbf{SFT (1.5B / 7B):} Qwen-1.5B and Qwen-7B backbones fine-tuned only with supervised oracle labels.
    \item \textbf{GRPO (1.5B / 7B):} The corresponding SFT-initialized Qwen models further optimized by GRPO without LACS ($\lambda=0$).
    \item \textbf{COMLLM:} The full proposed framework built upon the Qwen-7B backbone, using the same SFT initialization as GRPO-7B and further optimized by GRPO with the proposed LACS reward.
\end{itemize}

\subsection{Overall Performance Comparison}
Table~\ref{tab:performance_comparison} reports the overall performance of different offloading policies in the MEC environment with 11 edge servers. COMLLM achieves the best overall results, yielding the lowest average latency, a zero task drop rate, and the highest performance ratio relative to the one-step oracle reference. These results show that COMLLM can effectively optimize service quality while maintaining reliable deadline satisfaction.

\begin{table}[htbp]
    \centering
    \caption{Overall performance comparison of offloading policies}
    \label{tab:performance_comparison}
    \setlength{\tabcolsep}{3pt} 

    \begin{tabular}{lcccc}
        \toprule
        \multirow{2}{*}{\textbf{Model}}
        & \makecell{\textbf{Average}\\\textbf{Latency}}
        & \makecell{\textbf{Drop}\\\textbf{Rate}}
        & \makecell{\textbf{Performance}\\\textbf{Ratio}}
        & \makecell{\textbf{Load Balancing}\\\textbf{Index}} \\

        & \textbf{(time slots)}
        & \textbf{(\%)}
        & \textbf{(\%)}
        & \textbf{(\%)} \\
        \midrule
        COMLLM   & \textbf{3.0745} & \textbf{0.00} & \textbf{96.86} & \textbf{73.87} \\
        GRPO-7B  & 3.1197 & 0.00 & 95.46 & 71.20 \\
        DQN      & 3.3966 & 4.35 & 87.68 & 65.64 \\
        SFT-7B   & 4.0989 & 0.33 & 72.65 & 41.39 \\
        GRPO-1.5B & 4.3096 & 1.63 & 69.10 & 19.94 \\
        Random   & 4.5658 & 0.65 & 65.22 & 63.42 \\
        SFT-1.5B & 4.7441 & 2.94 & 62.77 & 58.67 \\
        \bottomrule
    \end{tabular}
\end{table}

Compared with COMLLM, GRPO-7B remains competitive but is consistently inferior across all metrics, indicating that the proposed LACS mechanism provides additional gains beyond standard RL fine-tuning. DQN performs noticeably worse, with higher latency and a non-negligible drop rate, suggesting that value-based DRL with fixed-dimensional numerical state representation is less effective in capturing long-term congestion effects in this dynamic MEC setting.

A clear performance gap is also observed between the 7B and 1.5B models. While the 7B-scale models achieve strong and stable results, the 1.5B variants degrade substantially and even fall below the random baseline in some metrics. This suggests that prompt-based MEC decision-making benefits not only from RL refinement, but also from sufficient model capacity to represent complex resource interactions. Overall, the results verify the effectiveness of combining semantic state representation, LLM-scale reasoning, and future-aware reward shaping in COMLLM.

\begin{table*}[htbp]
    \centering
    \caption{Performance under different task workload levels}
    \label{tab:data_size_comparison}

    \begin{tabular*}{\textwidth}
    {@{\extracolsep{\fill}} l cc cc cc cc cc @{}}
        \toprule
        \multirow{3}{*}{\textbf{Model}}
        & \multicolumn{2}{c}{\textbf{Task Size 2 Mbits}}
        & \multicolumn{2}{c}{\textbf{Task Size 4 Mbits}}
        & \multicolumn{2}{c}{\textbf{Task Size 6 Mbits}}
        & \multicolumn{2}{c}{\textbf{Task Size 8 Mbits}}
        & \multicolumn{2}{c}{\textbf{Task Size 10 Mbits}} \\

        \cmidrule(lr){2-3}
        \cmidrule(lr){4-5}
        \cmidrule(lr){6-7}
        \cmidrule(lr){8-9}
        \cmidrule(lr){10-11}

        & \makecell{\textbf{Average}\\\textbf{Latency}}
        & \makecell{\textbf{Drop}\\\textbf{Rate}}
        & \makecell{\textbf{Average}\\\textbf{Latency}}
        & \makecell{\textbf{Drop}\\\textbf{Rate}}
        & \makecell{\textbf{Average}\\\textbf{Latency}}
        & \makecell{\textbf{Drop}\\\textbf{Rate}}
        & \makecell{\textbf{Average}\\\textbf{Latency}}
        & \makecell{\textbf{Drop}\\\textbf{Rate}}
        & \makecell{\textbf{Average}\\\textbf{Latency}}
        & \makecell{\textbf{Drop}\\\textbf{Rate}} \\

        & \textbf{(time slots)}
        & \textbf{(\%)}
        & \textbf{(time slots)}
        & \textbf{(\%)}
        & \textbf{(time slots)}
        & \textbf{(\%)}
        & \textbf{(time slots)}
        & \textbf{(\%)}
        & \textbf{(time slots)}
        & \textbf{(\%)} \\
        \midrule
        
        \textbf{COMLLM}   & \textbf{1.8515} & \textbf{0.00} & \textbf{3.7409} & \textbf{0.00} & \textbf{5.3290} & \textbf{0.00} & \textbf{6.9759} & \textbf{1.08} & \textbf{9.5363} & \textbf{2.78} \\
        GRPO-7B   & 1.9723 & 0.00 & 3.8420 & 0.00 & 5.4595 & 0.00 & 7.1910 & 2.16  & 9.8527  & 4.31  \\
        SFT-7B    & 3.1702 & 0.00 & 4.9999 & 0.00 & 6.9515 & 3.03 & 9.2281 & 31.11 & 12.4634 & 55.02 \\
        GRPO-1.5B & 2.7558 & 0.00 & 5.1860 & 2.00 & 7.0620 & 8.08 & 8.5224 & 20.00 & 11.4648 & 28.23 \\
        SFT-1.5B  & 2.9676 & 1.01 & 5.2317 & 4.08 & 6.8924 & 5.15 & 9.1983 & 26.52 & 11.5179 & 30.81 \\
        \bottomrule
    \end{tabular*}
\end{table*}

\subsection{Robustness Under Extreme Task Workloads}
Table~\ref{tab:data_size_comparison} reports the performance of different methods under increasing task workloads in the default 6-server MEC environment. To stress the system near its capacity boundary, we gradually increase the task size while keeping the task deadline fixed. As the workload increases, all methods degrade, but the gap between COMLLM and the baselines becomes clear. In particular, under heavy workloads, imitation-based methods deteriorate rapidly, and even GRPO-7B exhibits an increase in task drop rate. By contrast, COMLLM achieves the lowest latency and the lowest task drop rate across all workload levels. These results suggest that the proposed LACS improves robustness under severe congestion by discouraging short-sighted offloading decisions.

\subsection{Topology Generalization}

\begin{table*}[htbp]
    \centering
    \caption{Performance under different edge-server topologies}
    \label{tab:server_scaling_performance}
    \setlength{\tabcolsep}{3pt} 
    \begin{tabular*}{\textwidth}{@{\extracolsep{\fill}} l ccc ccc ccc ccc ccc @{}}
        \toprule
        \multirow{3}{*}{\textbf{Model}}
        & \multicolumn{3}{c}{\textbf{3 Servers}}
        & \multicolumn{3}{c}{\textbf{5 Servers}}
        & \multicolumn{3}{c}{\textbf{7 Servers}}
        & \multicolumn{3}{c}{\textbf{9 Servers}}
        & \multicolumn{3}{c}{\textbf{11 Servers}} \\
        \cmidrule(lr){2-4}
        \cmidrule(lr){5-7}
        \cmidrule(lr){8-10}
        \cmidrule(lr){11-13}
        \cmidrule(lr){14-16}
        
        & \makecell{\textbf{Average}\\\textbf{Latency}}
        & \makecell{\textbf{Drop}\\\textbf{Rate}}
        & \makecell{\textbf{Perf.}\\\textbf{Ratio}}
        
        & \makecell{\textbf{Average}\\\textbf{Latency}}
        & \makecell{\textbf{Drop}\\\textbf{Rate}}
        & \makecell{\textbf{Perf.}\\\textbf{Ratio}}
        
        & \makecell{\textbf{Average}\\\textbf{Latency}}
        & \makecell{\textbf{Drop}\\\textbf{Rate}}
        & \makecell{\textbf{Perf.}\\\textbf{Ratio}}
        
        & \makecell{\textbf{Average}\\\textbf{Latency}}
        & \makecell{\textbf{Drop}\\\textbf{Rate}}
        & \makecell{\textbf{Perf.}\\\textbf{Ratio}}
        
        & \makecell{\textbf{Average}\\\textbf{Latency}}
        & \makecell{\textbf{Drop}\\\textbf{Rate}}
        & \makecell{\textbf{Perf.}\\\textbf{Ratio}} \\
        
        & \textbf{(time slots)}
        & \textbf{(\%)}
        & \textbf{(\%)}
        
        & \textbf{(time slots)}
        & \textbf{(\%)}
        & \textbf{(\%)}
        
        & \textbf{(time slots)}
        & \textbf{(\%)}
        & \textbf{(\%)}
        
        & \textbf{(time slots)}
        & \textbf{(\%)}
        & \textbf{(\%)}
        
        & \textbf{(time slots)}
        & \textbf{(\%)}
        & \textbf{(\%)} \\
        \midrule
        \textbf{COMLLM}    & \textbf{3.4097} & \textbf{0.00} & \textbf{93.90} & \textbf{3.0688} & \textbf{0.00} & \textbf{96.99} & \textbf{2.9026} & \textbf{0.00} & \textbf{98.59} & \textbf{2.8447} & \textbf{0.00} & \textbf{96.64} & \textbf{3.1114} & \textbf{0.00} & \textbf{97.42} \\
        GRPO-7B   & 3.4527 & 0.00 & 92.73 & 3.0718 & 0.00 & 96.90 & 2.9123 & 0.00 & 98.26 & 2.8518 & 0.00 & 96.41 & 3.1174 & 0.00 & 97.23 \\
        SFT-7B   & 4.2683 & 0.00 & 75.01 & 4.3316 & 0.00 & 68.72 & 3.5708 & 0.00 & 80.14 & 3.8498 & 0.00 & 71.41 & 4.4130 & 0.00 & 68.69 \\
        GRPO-1.5B & 4.6658 & 2.54 & 68.62 & 4.3707 & 0.96 & 68.10 & 4.3243 & 1.70 & 66.18 & 3.8503 & 0.00 & 71.40 & 4.7216 & 2.76 & 64.20 \\
        SFT-1.5B & 4.7812 & 3.05 & 66.96 & 4.4811 & 2.88 & 66.42 & 4.3252 & 1.04 & 66.17 & 4.3466 & 0.00 & 63.25 & 5.3499 & 6.21 & 56.66 \\
        \bottomrule
    \end{tabular*}
\end{table*}

Table~\ref{tab:server_scaling_performance} reports the performance of different methods under MEC topologies with varying numbers of edge servers. COMLLM was trained only under a 6-server topology and was directly evaluated on topologies with 3, 5, 7, 9, and 11 edge servers without topology-specific retraining or parameter updates. COMLLM consistently achieves the best overall performance across all tested topologies, maintaining the lowest latency, zero task drop rate, and the highest performance ratio. GRPO-7B remains competitive but is uniformly inferior to COMLLM, indicating that RL refinement alone improves adaptability, while the full COMLLM design provides stronger cross-topology robustness. In contrast, the SFT-only and 1.5B models degrade more noticeably as the topology changes, with higher latency, lower performance ratio, and nonzero drop rates in several settings. These results show that COMLLM generalizes well across different server topologies.

Although COMLLM achieves zero-shot transfer across the evaluated topologies, the present results do not establish unbounded scalability. Because semantic state serialization adds a structured description for each server, substantially larger MEC networks may introduce context-window limitations, attention dilution, and additional inference overhead. Extending COMLLM through multi-agent collaboration, where multiple agents process local network regions and coordinate their decisions, represents a promising direction for supporting larger-scale MEC systems.

\subsection{Load Balancing and Resource Allocation Fairness}
Table~\ref{tab:server_selection} shows the distribution of offloading decisions under the 11-server MEC topology. To quantify workload distribution across servers, we use Jain's Fairness Index as the load balancing metric. COMLLM achieves the highest fairness index, indicating the most balanced workload distribution among the compared methods. GRPO-7B shows a similar but slightly less balanced allocation pattern, while DQN and the SFT-based models exhibit more uneven server utilization. In particular, the imitation-based methods tend to over-concentrate decisions on a small subset of servers or fall back excessively to local execution, leading to noticeably lower fairness. These results suggest that COMLLM not only reduces latency, but also improves resource allocation fairness under a larger and more diverse action space.

\begin{table*}[htbp]
    \centering
    \caption{Offloading distribution and load balancing index}
    \label{tab:server_selection}
    \setlength{\tabcolsep}{4.5pt} 
    \begin{tabular}{l ccccccccccc c c}
        \toprule
        \textbf{Model} & \textbf{Local} & \textbf{Server1} & \textbf{Server2} & \textbf{Server3} & \textbf{Server4} & \textbf{Server5} & \textbf{Server6} & \textbf{Server7} & \textbf{Server8} & \textbf{Server9} & \textbf{Server10} & \textbf{Server11} & \makecell{\textbf{Load Bal.}\\\textbf{Index(\%)}} \\
        \midrule
        COMLLM   & 6   & 14  & 52  & 62  & 26  & 27  & 18  & 39  & 19  & 19  & 6   & 18  & \textbf{73.87} \\
        GRPO-7B   & 1   & 16  & 53  & 66  & 27  & 26  & 19  & 39  & 22  & 19  & 0   & 18  & 71.20 \\
        DQN       & 16  & 18  & 48  & 10  & 64  & 50  & 35  & 25  & 16  & 10  & 10  & 4   & 65.64 \\
        SFT-7B    & 136 & 9   & 67  & 30  & 17  & 4   & 11  & 1   & 0   & 3   & 9   & 19  & 41.39 \\
        GRPO-1.5B & 12  & 192 & 43  & 10  & 19  & 10  & 1   & 0   & 4   & 10  & 0   & 5   & 19.94 \\
        Random    & 59  & 46  & 42  & 51  & 26  & 31  & 16  & 15  & 10  & 6   & 1   & 3   & 63.42 \\
        SFT-1.5B  & 41  & 59  & 35  & 27  & 28  & 26  & 4   & 0   & 14  & 16  & 2   & 8   & 58.67 \\
        \bottomrule
    \end{tabular}
\end{table*}

\subsection{Prompt Robustness Under Semantic Perturbations}
To evaluate whether LLM-based policies rely on physically meaningful features rather than superficial prompt patterns, we test them under four prompt perturbations: Standard Prompt, Parameter Shuffling, Noisy Text Injection, and Unit Variation. Since this experiment focuses on text-conditioned decision policies, only LLM-based methods are compared. Table~\ref{tab:env_comparison} shows that COMLLM remains stable across all perturbation settings, achieving the best latency, zero task drop rate, and the highest load balancing index. GRPO-7B also maintains strong robustness, whereas the SFT-based models, especially SFT-1.5B, exhibit more noticeable performance variation under perturbed prompts. These results suggest that COMLLM learns a more robust semantic representation of the MEC state, enabling consistent decision-making even when the prompt format is altered or noisy.

\begin{table*}[htbp]
    \centering
    \caption{Robustness under semantic prompt perturbations}
    \label{tab:env_comparison}

    \setlength{\tabcolsep}{0pt} 
    
    \begin{tabular*}{\textwidth}
    {@{\extracolsep{\fill}} l ccc ccc ccc ccc @{}}
        \toprule
        \multirow{3}{*}{\textbf{Model}}
        & \multicolumn{3}{c}{\textbf{Standard Env.}}
        & \multicolumn{3}{c}{\textbf{Shuffled Params.}}
        & \multicolumn{3}{c}{\textbf{Noisy Env.}}
        & \multicolumn{3}{c}{\textbf{Unit Variation}} \\

        \cmidrule(lr){2-4}
        \cmidrule(lr){5-7}
        \cmidrule(lr){8-10}
        \cmidrule(lr){11-13}

        & \makecell{\textbf{Average}\\\textbf{Latency}}
        & \makecell{\textbf{Drop}\\\textbf{Rate}}
        & \makecell{\textbf{Load Bal.}\\\textbf{Index}}

        & \makecell{\textbf{Average}\\\textbf{Latency}}
        & \makecell{\textbf{Drop}\\\textbf{Rate}}
        & \makecell{\textbf{Load Bal.}\\\textbf{Index}}

        & \makecell{\textbf{Average}\\\textbf{Latency}}
        & \makecell{\textbf{Drop}\\\textbf{Rate}}
        & \makecell{\textbf{Load Bal.}\\\textbf{Index}}

        & \makecell{\textbf{Average}\\\textbf{Latency}}
        & \makecell{\textbf{Drop}\\\textbf{Rate}}
        & \makecell{\textbf{Load Bal.}\\\textbf{Index}} \\

        & \textbf{(time slots)}
        & \textbf{(\%)}
        & \textbf{(\%)}

        & \textbf{(time slots)}
        & \textbf{(\%)}
        & \textbf{(\%)}

        & \textbf{(time slots)}
        & \textbf{(\%)}
        & \textbf{(\%)}

        & \textbf{(time slots)}
        & \textbf{(\%)}
        & \textbf{(\%)} \\
        \midrule

        COMLLM   & \textbf{3.0675} & \textbf{0.00} & \textbf{94.71} & \textbf{3.0684} & \textbf{0.00} & \textbf{95.27} & \textbf{3.0617} & \textbf{0.00} & \textbf{94.18} & \textbf{3.0620} & \textbf{0.00} & \textbf{92.67} \\
        GRPO-7B   & 3.0722 & 0.00 & 93.20 & 3.0836 & 0.00 & 95.13 & 3.0868 & 0.00 & 93.12 & 3.0885 & 0.00 & 92.14 \\
        SFT-7B    & 4.3581 & 0.00 & 85.23 & 4.2393 & 0.00 & 87.76 & 4.1582 & 0.00 & 90.14 & 4.1999 & 0.00 & 84.94 \\
        GRPO-1.5B  & 4.3831 & 1.91 & 74.29 & 4.3666 & 1.47 & 69.07 & 4.3879 & 1.91 & 77.43 & 4.3710 & 0.48 & 69.21 \\
        SFT-1.5B  & 4.5025 & 2.44 & 41.21 & 4.4063 & 1.44 & 48.85 & 4.6933 & 4.78 & 59.40 & 4.4884 & 2.90 & 56.09 \\
        \bottomrule
    \end{tabular*}
\end{table*}

\section{Conclusion}
This paper investigates dynamic task offloading and long-term load balancing in heterogeneous MEC networks. We proposed \textbf{COMLLM}, a generative AI-driven framework that reformulates MEC offloading as a semantic reasoning task, addressing the rigidity of DRL and the short-sightedness of imitation-based LLM methods. By combining GRPO with the \textit{LACS} reward mechanism, COMLLM promotes more balanced long-term resource allocation. Simulation results show that COMLLM reduces latency and task drop rate while generalizing to unseen network topologies without retraining.

\bibliographystyle{IEEEtran} 

\bibliography{references} 

\end{document}


\appendices
\section{}

\subsection{Summary of Mathematical Quantities}
\label{app:notation_summary}
Section III formulates the communication delay, local and edge computation delay, server queue evolution, and sequential task offloading objective. For convenient reference, Table~\ref{tab:notation_summary} summarizes the principal variables and parameters appearing in Section III.

\subsection{Policy-Improvement Bound for GRPO}
\label{app:grpo_bound}
To analyze the effect of the GRPO policy update, define the discounted value function as
\begin{equation}
V(\pi)
=
\mathbb{E}_{\pi}
\left[
\sum_{t=0}^{\infty}\eta^t r_t
\right],
\end{equation}
where $\eta\in(0,1)$ is the discount factor.

\begin{theorem}[Performance Lower Bound for the Updated COMLLM Policy]
Let $\pi_{\theta_{\rm old}}$ denote the policy used to generate the group samples and let $\pi_\theta$ denote the updated policy. The performance of the updated policy satisfies
\begin{equation}
\begin{split}
V(\pi_\theta)
\geq V(\pi_{\theta_{\rm old}})
&+
\frac{1}{1-\eta}
\mathbb{E}_{
s\sim d^{\pi_{\theta_{\rm old}}},
\,a\sim\pi_\theta
}
\left[
A^{\pi_{\theta_{\rm old}}}(s,a)
\right]
\\
&-
\mathcal{B}
\max_s
\sqrt{
\mathbb{D}_{\rm KL}
\left(
\pi_\theta(\cdot|s)
\,\|\, 
\pi_{\theta_{\rm old}}(\cdot|s)
\right)
},
\end{split}
\label{thm:grpo_improvement}
\end{equation}
where
\begin{equation}
d^{\pi_{\theta_{\rm old}}}(s)
=
(1-\eta)
\sum_{t=0}^{\infty}
\eta^t
P(s_t=s\mid\pi_{\theta_{\rm old}})
\end{equation}
is the discounted state-visitation distribution under the old policy, $A^{\pi_{\theta_{\rm old}}}(s,a)$ is the corresponding advantage function, and
\begin{equation}
\mathcal{B}
=
\frac{
2\eta
\max_{s,a}
\left|
A^{\pi_{\theta_{\rm old}}}(s,a)
\right|
}{
(1-\eta)^2
}
\sqrt{\frac{1}{2}}
\end{equation}
is a positive constant.
\end{theorem}

\begin{proof}
The exact performance difference between any two policies is given by the policy-difference lemma:
\begin{equation}
V(\pi_\theta)-V(\pi_{\theta_{\rm old}})
=
\mathbb{E}_{\tau\sim\pi_\theta}
\left[
\sum_{t=0}^{\infty}
\eta^t
A^{\pi_{\theta_{\rm old}}}(s_t,a_t)
\right].
\end{equation}
Rewriting the trajectory expectation using the discounted state-visitation distribution gives
\begin{equation}
\begin{aligned}
V(\pi_\theta)-V(\pi_{\theta_{\rm old}})
=
\frac{1}{1-\eta}
\mathbb{E}_{
s\sim d^{\pi_\theta},
\,a\sim\pi_\theta
}
\left[
A^{\pi_{\theta_{\rm old}}}(s,a)
\right].
\end{aligned}
\end{equation}

Because $d^{\pi_\theta}$ depends on the updated policy, we introduce the surrogate
\begin{equation}
\begin{aligned}
L_{\pi_{\theta_{\rm old}}}(\pi_\theta)
=
V(\pi_{\theta_{\rm old}})
+
\frac{1}{1-\eta}
\mathbb{E}_{
s\sim d^{\pi_{\theta_{\rm old}}},
\,a\sim\pi_\theta
}
\left[
A^{\pi_{\theta_{\rm old}}}(s,a)
\right].
\end{aligned}
\end{equation}

The approximation error caused by replacing $d^{\pi_\theta}$ with $d^{\pi_{\theta_{\rm old}}}$ can be bounded using the total-variation distance:
\begin{equation}
\begin{aligned}
\left|
V(\pi_\theta)
-
L_{\pi_{\theta_{\rm old}}}(\pi_\theta)
\right|
&\leq
\frac{
2\eta
\max_{s,a}
\left|
A^{\pi_{\theta_{\rm old}}}(s,a)
\right|
}{
(1-\eta)^2
}
\\
&\quad\times
\max_s
D_{\rm TV}
\left(
\pi_\theta(\cdot|s)
\,\|\, 
\pi_{\theta_{\rm old}}(\cdot|s)
\right).
\end{aligned}
\end{equation}

Applying Pinsker's inequality,
\begin{equation}
D_{\rm TV}(P,Q)
\leq
\sqrt{
\frac{1}{2}
\mathbb{D}_{\rm KL}(P\|Q)
},
\end{equation}
yields
\begin{equation}
\begin{aligned}
V(\pi_\theta)
&\geq
L_{\pi_{\theta_{\rm old}}}(\pi_\theta)
\\
&\quad-
\mathcal{B}
\max_s
\sqrt{
\mathbb{D}_{\rm KL}
\left(
\pi_\theta(\cdot|s)
\,\|\, 
\pi_{\theta_{\rm old}}(\cdot|s)
\right)
}.
\end{aligned}
\end{equation}
Substituting the definition of
$L_{\pi_{\theta_{\rm old}}}(\pi_\theta)$ proves
\eqref{thm:grpo_improvement}.
\end{proof}

The bound indicates that the updated policy improves upon the previous policy whenever the expected advantage gain exceeds the divergence term. In GRPO, the clipped probability ratio stabilizes the estimated advantage contribution, while KL regularization constrains excessive policy deviation from the SFT-initialized reference policy.

\subsection{Proof of the LACS Mechanism's Bound}
\begin{proof}
Let $V^*(s)$ denote the true optimal expected cumulative cost from state $s$. Our LACS mechanism constructs a surrogate estimate of the downstream future cost, denoted by $\hat{V}(s)$, based on the expected simulation impact term, i.e., $\hat{V}(s)=\mathbb{E}[C_{\mathrm{impact}}(s_t,a_t)]$. Since we utilize Monte Carlo sampling to approximate the future traffic and an Oracle to evaluate it, there exists an estimation error $\epsilon_{sim}$.

To ensure the existence of a uniform constant simulation error bound \(\epsilon_{\rm sim} > 0\) for the Oracle value evaluation, we make the following standard modeling assumption that reflects both physical system constraints and theoretical requirements for a well-defined MDP. Assume the queue backlogs and task sizes in the MEC system are bounded, i.e., there exist finite constants \(L_{\max} < \infty\) and \(D_{\max} < \infty\) such that \(0 \leq L_e^t \leq L_{\max}\) for all edge servers \(e\) and time slots \(t\), and \(0 < D_m \leq D_{\max}\) for all sampled tasks \(m\). This assumption is consistent with practical hardware limitations and ensures that the relevant state variables used in the approximation argument remain bounded. The single-step Monte Carlo simulation error of the Oracle evaluation is bounded by a constant $\epsilon_{sim} > 0$, such that for any state $s$:
\begin{equation}
| \hat{V}(s) - V^*(s) | \leq \epsilon_{sim}\label{eq:simulation_error_bound}.
\end{equation}

Building upon the simulation error bound in Eq.~\eqref{eq:simulation_error_bound}, we establish the sub-optimality bound of the LACS-guided COMLLM policy compared to a purely greedy policy. Let \(\pi_{\rm greedy}\) be the policy that minimizes only the current-step cost \(J_t(a_t)\), and let \(\pi_{\rm LACS}\) be the COMLLM policy that optimizes the LACS-shaped reward.
By construction, the LACS-guided policy can be viewed as greedy with respect to the approximate action-value surrogate
\[
\tilde{Q}(s_t,a_t) = J_t(s_t,a_t) + \eta \hat{V}(\hat{s}_{t+1}),
\]
From Eq.~(\ref{eq:simulation_error_bound}), the maximum error of the future-value estimate satisfies \(\|\hat{V} - V^*\|_\infty \leq \epsilon_{\rm sim}\). According to standard approximation bounds for discounted dynamic programming with bounded state variables, if a policy \(\pi\) is greedy with respect to an approximate value function \(\hat{V}\) whose maximum error is bounded by \(\epsilon_{\rm sim}\), then the worst-case performance loss of \(\pi\) relative to the optimal value function \(V^*\) satisfies
\begin{equation}
\|V^* - V^\pi\|_\infty \leq \frac{2\eta \|\hat{V} - V^*\|_\infty}{1-\eta} \leq \frac{2\eta \epsilon_{\rm sim}}{1-\eta}.
\label{eq:greedy_approx_bound}
\end{equation}
Consequently, for the LACS-guided policy \(\pi_{\rm LACS}\), the sub-optimality gap is bounded by
\begin{equation}
V^{\pi_{\rm LACS}}(s) -V^*(s) \leq \frac{2\eta \epsilon_{\rm sim}}{1-\eta}
\label{eq:lacs_suboptimality}
\end{equation}
for any state \(s\). This result suggests that, when the simulation error remains sufficiently small, the LACS-guided policy admits a tighter approximation of downstream future cost than a purely myopic greedy policy that ignores congestion evolution. In this sense, the bound provides theoretical support for introducing LACS into the reward design, since it shows that incorporating a simulation-based future-impact term can reduce the suboptimality induced by short-sighted decisions.
\end{proof}

\begin{table*}[!t]
\centering
\caption{Principal Mathematical Quantities Appearing in Section III}
\label{tab:notation_summary}
\renewcommand{\arraystretch}{1.12}
\setlength{\tabcolsep}{4pt}
\scriptsize
\begin{tabular}{
>{\centering\arraybackslash}p{0.12\textwidth}
p{0.35\textwidth}
>{\centering\arraybackslash}p{0.12\textwidth}
p{0.35\textwidth}}
\toprule
\textbf{Symbol} & \textbf{Meaning}
& \textbf{Symbol} & \textbf{Meaning} \\
\midrule

$R_{u,e}^{\mathrm{up}}$
& Uplink transmission rate from user $u$ to edge server $e$
& $B$
& Wireless channel bandwidth \\

$P$
& Transmission power of the mobile user
& $|h_{u,e}|^2$
& Channel power gain between user $u$ and edge server $e$ \\

$\sigma^2$
& Noise power
& $T_m^{\mathrm{up}}$
& Uplink transmission delay of task $m$ \\

$x_{m,e}$
& Binary variable indicating whether task $m$ is assigned to execution location $e$
& $D_m$
& Input data size of task $m$ \\

$\gamma$
& Computational density in CPU cycles per bit
& $f^{\mathrm{loc}}$
& Local CPU frequency \\

$T_{m,0}^{\mathrm{wait,loc}}$
& Local queue waiting delay of task $m$
& $T_{m,0}^{\mathrm{loc}}$
& Total local execution latency of task $m$ \\

$F_e$
& Maximum computation capacity of edge server $e$
& $N_e^m$
& Number of active tasks at edge server $e$ when task $m$ arrives \\

$f_{m,e}^{\mathrm{eff}}$
& Effective computation rate allocated to task $m$ at edge server $e$
& $L_e^m$
& Workload backlog of edge server $e$ when task $m$ arrives \\

$T_{m,e}^{\mathrm{exec}}$
& Execution delay of task $m$ at edge server $e$
& $T_{m,e}^{\mathrm{wait}}$
& Queueing delay of task $m$ at edge server $e$ \\

$T_{m,e}^{\mathrm{edge}}$
& Total latency when task $m$ is offloaded to edge server $e$
& $T_{m,e}$
& Candidate execution latency of task $m$ at location $e$ \\

$R_{m,e}^{\mathrm{process}}$
& Equivalent input workload processed by edge server $e$ in one time slot
& $\Delta t$
& Duration of one time slot \\

$L_e^{m+1}$
& Updated workload backlog of edge server $e$
& $C_m$
& Latency resulting from the selected execution location \\

$\tau_m^{\max}$
& Maximum tolerable latency of task $m$
& $\rho$
& Penalty for violating the task deadline \\

$J_m$
& Generalized task cost including the deadline-violation penalty
& $\pi$
& Stochastic task-offloading policy \\

\bottomrule
\end{tabular}
\end{table*}

\subsection{Prompt Template and Design Rationale}
\label{app:prompt_template}

To improve the reproducibility of the semantic state serialization, we provide the complete prompt template used for COMLLM training. The following template is an English translation of the original Chinese prompt and is written using the notation adopted in the main paper. The terms enclosed in braces are instantiated using the current MEC state. The edge-server description is repeated for every currently available server $e\in\mathcal{E}$.

The prompt is designed according to four principles. First, all numerical variables are paired with semantic labels and physical units, which helps distinguish heterogeneous quantities such as task size, computation capacity, transmission rate, and queue backlog. Second, the task, local device, and edge-server information are organized into separate blocks so that their respective physical roles remain explicit. Third, every edge server follows the same field order and key--value structure. The server block is repeated according to the current topology; therefore, adding or removing a server changes only the prompt length and the feasible action list, without requiring padding, architectural modification, or retraining. Fourth, the previous 10-slot active-task sequence provides short-term temporal information about the server-load trend, complementing the instantaneous active-task number and queue backlog. The instruction that the model return only the decision name also constrains the output format and enables deterministic mapping from the generated response to the discrete action space $\mathcal{A}$. Therefore, the prompt jointly provides semantic grounding, topology flexibility, temporal context, and a constrained decision interface.

\subsection{Training Setup and Computational Cost}
\label{app:training_setup}

The SFT stage was conducted on one NVIDIA GeForce RTX 3060 GPU with 6 GB memory using LoRA-based parameter-efficient fine-tuning. LoRA was applied to the attention and feed-forward projection modules, with rank $r=64$, scaling parameter $\alpha=128$, and dropout rate $0.05$. The model was trained for one epoch using BF16 precision, a per-device batch size of 2, four gradient accumulation steps, a learning rate of $1\times10^{-5}$, and a maximum sequence length of 2,048 tokens. The effective batch size was therefore 8 samples.

The subsequent GRPO stage was conducted on one NVIDIA GeForce RTX 5090 GPU with 32 GB memory. It was trained for one epoch using BF16 precision, a per-device prompt batch size of 1, four gradient accumulation steps, a learning rate of $5\times10^{-6}$, and a KL coefficient of $0.005$. For each prompt, GRPO generates $G=8$ candidate responses, with a maximum completion length of 15 tokens. In the full COMLLM framework, LACS further samples $K=3$ future tasks for each candidate and evaluates their costs under the feasible actions. Therefore, GRPO and LACS increase the offline training cost relative to SFT. During deployment, however, only a single policy inference is performed, without grouped generation, policy updates, or LACS simulation.

\begin{figure*}[!t]
\centering
\begin{tcblisting}{
    colback=gray!5,                
    colframe=black!60,             
    boxrule=0.5pt,                 
    arc=3pt,                       
    left=3pt, right=3pt, top=3pt, bottom=3pt,
    listing only,                  
    listing options={
        basicstyle=\ttfamily\footnotesize,
        breaklines=true,
        columns=flexible,      
        keepspaces=true,      
        mathescape=true,
        breakindent=2.5em,     
        breakautoindent=true   
    },
    width=0.96\textwidth           
}
{
  "instruction":
  "You are an MEC task-offloading decision agent. Your objective is to determine an offloading decision for each newly arrived task so as to minimize its total processing latency while avoiding task dropping. A task is dropped when its total latency exceeds the maximum tolerable latency, resulting in a predefined penalty.

  Decision procedure:
  Evaluate the estimated latency of each feasible action, including local execution and offloading to every available edge server.

  1. Local execution latency consists of the local CPU queue waiting time and the local computation time. The computation time depends on the task size, computational density, and local CPU frequency.

  2. Edge execution latency consists of the uplink transmission time, edge-server queue waiting time, and edge-server execution time. These terms depend on the task size, uplink rate, server backlog, computational density, CPU capacity, and number of active tasks. When evaluating a new offloaded task, the number of active tasks is increased by one.

  3. The active-task history over the previous 10 time slots is used to describe the recent server-load trend.

  Based on the following state, select the feasible action with the lowest estimated total latency while considering the task deadline. Return only a JSON object containing the selected decision name in the specified output format.",

  "input": {
    "Task size": "{$D_t$} Mbits",
    "Maximum tolerable latency": "{$\tau_t^{\max}$} time slots",
    "Local CPU queue waiting time": "{$T_{\mathrm{wait}}^{\mathrm{loc}}$} time slots",
    "Time-slot duration": "{$\Delta_t$} s",
    "Task-drop penalty": "{$\rho$}",

    "Local device": "CPU frequency {$f_{\mathrm{loc}}$} GHz, computational density {$\gamma$} cycles/bit",

    "Edge server 1": "CPU frequency {$F_1$} GHz, computational density {$\gamma$} cycles/bit, uplink rate {$R_{\mathrm{up,1}}$} Mbps, active tasks {$N_1$}, queue backlog {$L_1$} Mbits, previous 10-slot load history {$H_1$}",

    "Edge server 2": "CPU frequency {$F_2$} GHz, computational density {$\gamma$} cycles/bit, uplink rate {$R_{\mathrm{up,2}}$} Mbps, active tasks {$N_2$}, queue backlog {$L_2$} Mbits, previous 10-slot load history {$H_2$}",

    "...": "...",

    "Edge server E": "CPU frequency {$F_E$} GHz, computational density {$\gamma$} cycles/bit, uplink rate {$R_{\mathrm{up,E}}$} Mbps, active tasks {$N_E$}, queue backlog {$L_E$} Mbits, previous 10-slot load history {$H_E$}"
  },

  "output": {
    "action_name": "<Local execution | Offload to Edge Server e>"
  }
}
\end{tcblisting}
\vspace{-2mm}
\caption{Complete semantic prompt template used for MEC task-offloading decision generation.}
\label{fig:prompt_template}
\end{figure*}